\documentclass[a4paper]{article}

\usepackage{INTERSPEECH2021}
\usepackage[hidelinks]{hyperref}
\usepackage{url}

\title{A Study of Different Ways to Use The Conformer Model For Spoken Language Understanding}
\name{Nick J.C. Wang, Shaojun Wang, Jing Xiao}
\address{
  Ping An Technology}
\email{}

\begin{document}

\maketitle
\begin{abstract}
SLU combines ASR and NLU capabilities to accomplish speech-to-intent
understanding. In this paper, we compare different ways to combine ASR 
and NLU, in particular using a single Conformer model with
different ways to use its components, to better understand the
strengths and weaknesses of each approach. We find that it is not
necessarily a choice between two-stage decoding and end-to-end systems
  which determines the best system for research or application.  
System optimization still entails
carefully improving the performance of each component. It is difficult to 
  prove    
that one direction is 
conclusively    
better than the other. In this paper,
we also propose a novel connectionist temporal summarization (CTS) method to
reduce the length of acoustic encoding sequences while improving the
accuracy and processing speed of end-to-end models. This
method achieves the same intent accuracy as the best
two-stage SLU recognition with complicated and time-consuming decoding but does so at lower
computational cost. This stacked end-to-end SLU system yields an intent accuracy of
93.97\% for the SmartLights far-field set, 95.18\% for the close-field set, and 
99.71\% for FluentSpeech.
\end{abstract}

\section{Introduction}

There has been a recent rise in the study of end-to-end SLU methods
~\cite{Qian2017E2ESLU,Serdyuk2018E2ESLU,Chen2018E2ESLU,Lugosch2019,NickWang2021SLU,McKenna2020SemanticCI,PengweiWang2020,price2020end,Price2020ImprovedES,price2020end,Rongali2021ExploringTL,Jia2020LargescaleTL,Lugosch2020UsingSS}
,
which have a common goal of finding more accurate 
and rapid approaches to solving the problem of recognizing 
the intent in an utterance directly from the voice 
without using ASR to produce the text.
One of the primary difficulties of NLU is ASR errors;  
thus the lack of ASR textual output 
is likely to reduce certain errors in intent recognition.
However, it is yet uncertain whether end-to-end approaches are more effective 
in reducing errors.
Therefore, we compare the end-to-end approach 
with the traditional two-stage approach.
In particular, to account for the differences 
from different ASR components,
in the experiments here we use a single ASR model.
After controlling for ASR variables, 
we seek to understand the differences 
in accuracy and speed that arise from differences in system design.

The complexity of sentence expressions 
in SLU application scenarios varies greatly.
These include simple command and control scenarios, 
where some include fixed expressions only and some are more open.
SLU recognition with fixed expressions is easier, 
but dialogue systems are typically characterized by more complex utterances.
Not only are there no fixed expressions, 
but sentence lengths vary greatly, 
and the fluency and speed of speech vary, 
making overall SLU recognition more difficult.
However, the recognition ability of neural networks 
depends heavily on the training corpus. 
Different setups of test data against training data 
may also highly influence to the evaluation results,
especially in the SLU systems that are composed of multiple  sub-task components
~\cite{Arora2021RethinkingEE}.
Different setups of test data against training data 
may also highly influence to the evaluation results,
especially in the SLU systems that are composed of multiple  sub-task components
~\cite{Arora2021RethinkingEE}.
For simple SLU applications, it is easier to collect training speech
that covers all expressions with rich pronunciation variations
among different speakers. 
In contrast, it is difficult to collect data for complex scenarios, 
or even to collect a sufficient number of sentence patterns.
For simple scenarios, neural networks can be trained 
  as data is sufficient:   
a simple type of neural network may fit the data and work quickly.
For complex scenarios where there is not enough training data, 
neural networks are needed 
that provide more accurate language prediction capability 
in situations where there is not enough data.
The design of a good SLU system must account for all of these issues.
Different to \cite{Arora2021RethinkingEE} designing different challenge test sets with held-out speaker or utterance for precise evaluation, we in this paper use the same ASR component to reduce biases in comparing different SLU combination methods.  

Current end-to-end studies in the literature show that 
SLU models trained using multi-objective learning 
that includes text and phoneme annotations 
outperform single models that target intent alone
~\cite{Chen2018E2ESLU,Lugosch2019,NickWang2021SLU,McKenna2020SemanticCI,PengweiWang2020}
.
That is, using ASR annotations to compute loss 
and to perform error back-propagation learning 
for the associated neural network parameters 
still improves the intent recognition of end-to-end SLU.
However, current end-to-end SLU models
still use a simple stack of ASR models with NLU,
and most of these stacked models are RNNs.
In contrast, we use Conformer as the ASR model
~\cite{Gulati2020ConformerCT} 
Conformer uses self-attention instead of RNN.
In addition, 
in order to perform hybrid-loss training,
Conformer contains CTC- and attention-based decoders, 
which increases the richness of our comparisons 
with two different ASR decoding methods.  
CTC decoding is based on conditional independent probability, 
whereas attention decoding is based on conditional dependent probability.
The latter might be useful as applying on SLU 
with insufficient training utterances to cover all testing ones.

From an ASR perspective, 
the Conformer model structure is currently
  state-of-the-art.                      
As already mentioned,
it effectively utilizes the hybrid loss of CTC and attention 
to train the model parameters.
CTC, the first end-to-end technique developed for ASR~\cite{Graves2006,Graves2014},
automatically aligns sounds to words via dynamic programming 
and requires no pre-given alignment.
Compared with the cross-entropy (CE) training for a given path,
it enriches the neural network training with different alignment paths
based on forward-backward learning. 
Different paths represent different ways to pronounce the same word label sequence;
this improves the robustness of the model.
However, CTC is still limited by its assumption of conditional independence,
and it makes poor use of rich linguistic knowledge. 
The multi-head encoder-decoder-attention mechanism, on the other hand,
incorporates rich sentence-wide attention~\cite{Vaswani2017,Dong2018SpeechTransformerAN,Vila2018EndtoEndST}. 
CTC/attention multi-target learning (MTL) 
further combines the strengths of CTC and attention~\cite{Watanabe2017, Miao2019}. 
The Conformer design includes all of these advances
and uses convolution blocks in the encoder to 
improve its robustness to speech dynamics~\cite{Gulati2020ConformerCT}.

Below, we describe how this study utilizes the various Conformer components to
construct different SLU systems.

\section{Research Methods}

In all of the experiments in this paper, 
we use Conformer as the ASR model.
This uses a hybrid CTC/attention loss
to train the model parameters,
which yields high accuracy and robust recognition capability.
Since it adopts both CTC and attention decoding, 
we take advantage of both to put together two two-stage SLU systems.
In addition, the CTC decoding method only uses the encoder component, 
which facilitates the comparison of stacked end-to-end systems
by using this encoder to form a cascaded end-to-end SLU model and ignore
ASR decoding.
These two systems differ only in their ASR decoding. 
This comparison thus affords a fair view into the difference 
between pipelined and end-to-end systems without ASR biases.  
Additional details are provided later.

\begin{figure}[t]
  \centering
  \includegraphics[width=\linewidth]{
  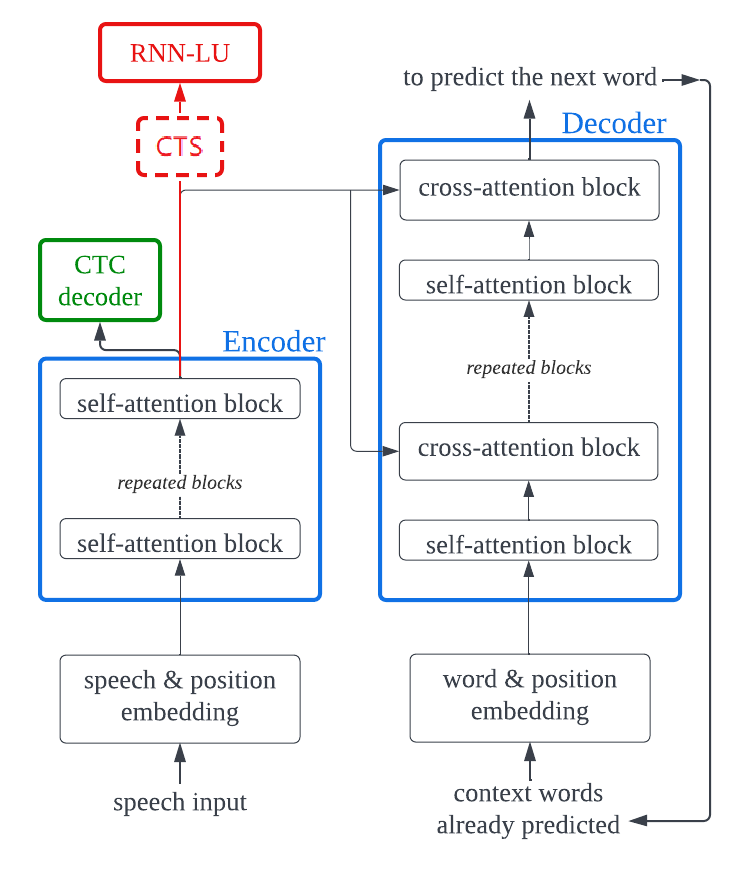}
  \caption{Conformer structure and various output types to two types
  as an end-to-end LU processing system}
  \label{fig_conformer}
\end{figure}

To construct each system, we take the following steps.
First, for each dataset the Conformer model is trained and evaluated for ASR. 
Section~\ref{sec_asr} compares the accuracy of the two decoding methods on each dataset.
Then, 
the Conformer encoder works as an acoustic encoding unit to convert speech from a
sequence of Mel-frequency filter bank coefficients, $\mathbf{X}  =
 \{\mathbf{x}_1 \dots \mathbf{x}_T\}$,    
to a sequence of acoustic latent representations, $\mathbf{Y} = \mathop{\mbox{Enc}}(\mathbf{X})$,  
where $T$ denotes the number of frames. 
Sequence $\mathbf{Y}$
is then decoded via the CTC algorithm to a sequence of word-pieces, $\hat{W}_{\mathit{CTC}}$, as 
\begin{equation}
  \hat{W}_{\mathit{CTC}} = \mathop{\arg\max} \limits_{W \in \mathcal{V}} 
  \mathop{\mbox{CTC}}(\mathop{\mbox{Enc}}(\mathbf{X}) | {\theta_{\mathit{CTC}}}),
  \label{eq_ctc}
\end{equation}
after which comes the second stage of word NLU recognition, described in Section~\ref{sec_P1}.
The Conformer encoder is ideal for acoustic coding of speech:                         
its outcome sequence represents a set of acoustic                                     %
feature vectors that can be used to recognize words                                   %
or even to recognize the embedded intent carried in these words.                      %
Therefore, 
a stacked end-to-end SLU approach could
feed these vectors as input to an NLU module 
to classify the intent directly, skipping ASR decoding,
as in the system described in Section~\ref{sec_E1}.
In addition, we propose a
connectionist time summary (CTS) component 
to shorten the acoustic sequence 
before sending it to the RNN LU module in end-to-end methods, 
as described in Section~\ref{sec_E2}.
The last comparison is of the pipeline approach using Conformer's
attention-rescoring decoding. Equation~\ref{eq_att} predicts $\hat{W}_{\mathit{Att}}$ 
using both the encoder and the decoder:
\begin{equation}
  \hat{W}_{\mathit{Att}} = \mathop{\arg\max} \limits_{W \in  \mathcal{V}} 
  \mathop{\mbox{Att}}(\mbox{Enc}(\mathbf{X}) | W^h, {\theta_{\mathit{Dec}}}).
  \label{eq_att}
\end{equation}
The attention mechanism utilizes a conditional dependent distribution which
treats previously decoded labels as the condition when predicting the next
word. 
If it produces better ASR output than CTC decoding, 
we expect that it would form a better pipeline system, as described in
Section~\ref{sec_P2}. However, computational cost may be a concern, 
as when using a larger beamwidth,    
as discussed in Section~\ref{sec_cpu}.
The various ways of using the Conformer components 
are illustrated in Fig.~\ref{fig_conformer}.

\subsection{Pipeline P1: Conformer encoder, CTC decoding, and text-based NLU}
\label{sec_P1}

P1 is a pipelined system that accomplishes ASR decoding using
Conformer's encoder and a CTC component, depicted as `CTC decoder'
in the upper-left of Fig.~\ref{fig_conformer}. 
The ASR decoding outcome, $\hat{W}_{\mathit{CTC}}$, is fed into a text-based NLU
model to produce the final SLU result, $\mathop{\mbox{NLU}}(\hat{W}_{\mathit{CTC}})$. 

Conformer is pretrained with LibriSpeech and fine-tuned with target SLU
data, either FluentSpeech or SmartLights. The text-based NLU network is trained
with the SLU data only. 

\subsection{End-to-end E1: stacked Conformer encoder and RNN LU}
\label{sec_E1}

E1 is an end-to-end SLU system that skips ASR decoding
and produces latent sequences, 
  $\mathbf{Y} = \{\mathbf{y}_t | \mathop{\mbox{Enc}}(\mathbf{x}_t), t = 1 \dots T\}$,   
from the ASR module.
These sequences are fed directly to an LU module, the
RNN-LU in the upper-left of Fig.~\ref{fig_conformer}, to
classify the intent, $\mathop{\mbox{LU}}(\mathbf{Y})$. 

Conformer is used exactly the same as in P1.            
However, the RNN-LU module
is trained with the SLU data 
  processed by the Conformer encoder.              

\subsection{End-to-end E2: stacked Conformer encoder, connectionist
temporal summarization (CTS), and RNN LU}
\label{sec_E2}

In the third system, we add a component to the well-trained E1 system:
connectionist temporal summarization (CTS).
Its function is to shorten the sequence $\mathbf{Y}$ 
to a shorter one, $\mathbf{Y'} = \mathop{\mbox{CTS}}(\mathbf{Y})$,
to reduce the RNN-LU computational complexity.
NLU is then fed with the resulting sequence, $\mathop{\mbox{LU}}(\mathbf{Y'})$. 
The algorithm is described in detail in Section~\ref{sec_cts}.
Since the input to NLU has changed, 
a fine-tuning step for RNN-LU is necessary. 
This is done using the pre-trained parameters from E1.

\subsubsection{Connectionist temporal summarization (CTS)}
\label{sec_cts}

CTS processing shortens the latent sequence $\mathbf{Y}$ from
frame-wise to segment-wise.  
We compute the softmax distributions for each frame $\mathbf{y}_t$ in the
acoustic encoding sequence
with the help of the linear transformation matrix $H_{\mathit{CTC}}$ in the CTC-loss component.
This maps 
  $\mathbf{y}_t$    
from the dimension of the latent representation vector for each frame
to the size of the word-piece labels in vocabulary $\mathcal{V}$ as 
\begin{equation}
  p_{w,t} = \mathop{\mbox{softmax}}\limits_{w \in \mathcal{V}}(H_{\mathit{CTC}} \cdot 
    \mathbf{y}_t).    
  \label{eq_softmax}
\end{equation}
We then determine the label with the maximal softmax score for each
frame using 
\begin{equation}
  \hat{w}_t = \mathop{\arg\max}\limits_{w \in \mathcal{V}}(p_{w,t}),
  \label{eq_argmax}
\end{equation}
after which we perform the following two steps: 
\begin{itemize}
\item {\bf{Segmentation}}: we aggregate as one segment all contiguous frames 
with the same maximum label $\hat{w}_{t_s} = \cdots = \hat{w}_{t_e}$.
\item {\bf{Representation}}: for each segment, we find the frame $\hat{t}$ with
the maximal softmax score via Eq.~(\ref{eq_repre})
	 and select its vector $y_{\hat{t}}$ to represent the whole segment.
\end{itemize}
\begin{equation}
  \hat{t} = \mathop{\arg\max}\limits_{{t_s} \leq t \leq {t_s}}(p_{\hat{w},t})
  \label{eq_repre}
\end{equation}
Note that the CTS shortening 
resembles the max-pooling effect over time for its segmental
selection and representation processing 
as illustrated in Eq.~\ref{eq_maxpool}:
\begin{equation}
  \mathbf{Y'} = 
  \mathop{\mbox{CTS}}(\mathbf{Y})
  \cong
  \{ \mathbf{y'}_i | \mathop{\mbox{maxpool}}\limits_{{t_{s_i}} \leq t \leq {t_{s_i}}}(\mathbf{y}_t), i=1 \dots S \} , 
\label{eq_maxpool}
\end{equation}
where $i$ is the segment index. 
After shortening, the CTS sequence lengths are usually twice as long as the   
number of word-pieces in each utterance, 
which may help LU to be more efficient than when using
acoustic sequences as input. 

\subsection{Pipeline P2: Conformer encoder-decoder,
attention-rescoring decoding, and text-based NLU}
\label{sec_P2}

The fourth system is a pipeline system that uses Conformer's 
attention-rescoring decoding.
Compared with CTC decoding, this requires at least doubles the running time.
Its larger beam widths dramatically increase its time consumption. 
It uses the previous decoded labels as conditions.
As in P1,           
its decoding outcome, $\hat{W}_{\mathit{Att}}$, 
is fed to a text-based NLU model to decide the final intent decision, 
$\mathop{\mbox{NLU}}(\hat{W}_{\mathit{Att}})$. 
Both Conformer and the text-based NLU network are exactly as the same as in P1: 
only ASR decoding is different.

\section{Experiments}

\subsection{Datasets}

In our experiments we used the open-source Librispeech 
ASR corpus~\cite{Panayotov2015}\footnote{\url{http://www.openslr.org/12}} as the primary
dataset when training Conformer. 
This is a corpus of read speech derived from audiobooks, comprising approximately 1000 hours
of 16\,kHz read English speech. 
We used the 960-hour training set of clean and noisy speech,
the 5.4-hour dev-clean set for validation, and the 5.4-hour test-clean
set for testing. 

Fluent Speech Commands,\footnote{\url{https://fluent.ai/research/fluent-speech-commands/}}  
or FluentSpeech, is a corpus
of SLU data for spoken commands to smart homes or virtual assistants,
for instance, ``put on the music'' or ``turn up the heat in the kitchen''.
Each audio utterance is labeled with three slots: \emph{action}, \emph{object}, and \emph{location}. 
A slot takes one of multiple values: for instance, \emph{location} can
take the values ``none'', ``kitchen'', ``bedroom'', or ``washroom''. 
Here, we follow Lugosch et~al.\ in referring to the
combination of slot values as the intent of the utterance, without
distinguishing between domain, intent, and slot prediction~\cite{Lugosch2019}.
The dataset contains 31 unique intents.

Snips SmartLights~\cite{Smartlights}, another corpus of SLU data for spoken
commands to smart homes, 
contains both close-field and far-field acoustic recording conditions. 
The latter was recorded via the replay-and-record approach with a microphone at
a distance of 2~meters. 
The dataset contains 6 unique intent classes. 
Because SmartLights does not define training and test sets, we selected the
first, second, third, fourth, and fifth entries of every five utterances as
five different training and evaluation sets for our five-fold cross
validation experiments. 

\begin{table}[thb]
  \caption{Speech datasets}
  \label{tab:datasets}
  \centering
  \begin{tabular}{lr|rrr}
    \hline 
    \textbf{Dataset} & \hspace{-5mm}\textbf{Intents} & \hspace{-5mm}\textbf{Spkrs} & \textbf{Utts} & \hspace{-2mm}\textbf{Hours}                \\
    \hline 
    LibriSpeech & & &  &     \\
    \hspace{5mm} train      & -- & 2,338  & 281,241     & 960.7       \\
    \hspace{5mm}  dev-clean & --    & 97    & 2,703     & 5.4       \\
    \hspace{5mm}  test-clean & --     & 87    & 2,620     & 5.4       \\
    FluentSpeech & & &  &     \\
    \hspace{5mm} train  & 31    & 77    & 23,132    & 14.7      \\
    \hspace{5mm} valid  & 31    & 10    & 3,118     & 1.9       \\
    \hspace{5mm} test   & 31    & 10    & 3,789     & 2.4       \\
    SmartLights & & & & \\
    \hspace{5mm} close-field & 6 & 52  & 1,660  & 1.4 \\
    \hspace{5mm} far-field   & 6 & 52  &  1,660 & 1.4 \\
    \hline 
  \end{tabular}
\end{table}

\subsection{ASR accuracy}
\label{sec_asr}

We started with ASR training and testing. The Conformer model was pre-trained
with LibriSpeech and fine-tuned with SLU data. 
All models were built using the ESPNet 
toolkit\footnote{\url{https://github.com/espnet/espnet}} using Mel-frequency
filter bank (Fbank) feature vectors as input; 
this is a sequence of 83-dimensional feature vectors 
with a 25-ms window size and 10-ms window shifts. 
The model used 2-layer convolutional neural networks (CNN) as the
frontend, each of which had 256 filters with 3$\times$3 kernel size
and 2$\times$2 stride; thus the time reduction of the frontend was 1$/$4.
Our vocabulary was a list of 5000 byte-pair-encoded~\cite{shibata1999byte} word-piece labels.
The model contained 12~layers of self-attention encoder blocks and 6~layers
of decoder blocks, each of which contained a self-attention layer
and an encoder-decoder cross-attention layer. 
The multi-head attentions in both models had
four heads, 256 attention dimensions, and 1024-dimensional feedforward networking.
They were trained with hybrid CTC/attention loss with a CTC weight of~0.5.
As in the decoding stages, 0.3 was used instead.

We evaluated the ASR accuracies with both CTC decoding and attention-rescoring
decoding on each test set. The results are listed in Table~\ref{tab:asr}. As
mentioned above, the SmartLights experiments were conducted using 5-fold cross
validation. Conformer yielded good accuracy for SmartLights even with
the few and sparse data available for training. 

\begin{table}[ht]
  \caption{Word error rate (WER \%) evaluation of ASR using CTC decoding and
  attention-rescoring decoding on different datasets}
  \label{tab:asr}
  \centering
  \begin{tabular}{lccc}
    \toprule
    Decoding  	& \multicolumn{1}{c}{\bf{CTC}} & \multicolumn{2}{c}{\bf{Att-rescoring}} \\
    Beamwidth & no & 1 & 4 \\
    \midrule
    {\bf LibriSpeech} 	& 4.04 & 4.75 & \bf{3.78} \\
    {\bf FluentSpeech}	& 1.18 & \bf{1.09} & 1.10 \\
    {\bf SmartLights} close-field & 5.75 & \bf{5.66} & 5.78 \\
    {\bf SmartLights} far-field & 7.32 & 7.14 & \bf{7.09} \\
    \bottomrule
  \end{tabular}
\end{table}

\subsection{SLU accuracy}
\label{sec_slu}

We built RNN LU modules for systems~E1 and E2 using the acoustic
latent-sequence input from the Conformer encoder, instead of using text labels. 
The end-to-end RNN LU networks and the text-based NLU networks were identical in
their sizes and types. 
We used a two-layer bidirectional LSTM network with 256~neurons per layer and
per direction. 
E2 used the pre-trained NLU network inherited from E1, 
as CTS sequences were summarized from those sequences
used to train the NLU network in E1. 
All end-to-end models were trained using PyTorch 
by adding the NLU network on top of the Conformer encoder. 
In our experiments, as 
with the  
training NLU parameters,
the parameters for the Conformer encoder were all fixed without change. 

The intent classification accuracy of all compared SLU systems is reported in
Table~\ref{tab:slu}. 
As mentioned above, the SmartLights experiments used 5-fold cross
validation. As these used the Conformer for ASR, all SLU accuracies were high. 
The best systems were the P2 pipeline system using attention-rescoring ASR decoding and 
the E2 end-to-end system using CTS sequences. 
Although E2 was derived from E1 by shortening acoustic encoding sequences, 
it yielded even better accuracy with less input data. 
Also, note that while attention-rescoring ASR decoding makes use of sentence-wide word-association
probabilities, the E2 end-to-end system does not. 
This supports the use of the CTS encoding sequence as a compact representation form 
in end-to-end SLU applications. 

As for the comparison between pipeline and end-to-end 
methods, it is unclear which is better. 
While end-to-end E1 is outperformed by pipeline P1, 
E2 is better than P1. 
Everything is important in each recognition step 
over the whole SLU process, 
including the input representation ability,
the neural network capability, 
and the training approach. 
Whether we proceed from speech to textual words 
or to latent acoustic sequences, 
and whether we go from 
textual words or from latent acoustic sequences 
to intent,
every step counts. 

\begin{table}[ht]
  \caption{SLU intent classification accuracy (\%) using CTC
  decoding and full Conformer decoding on different datasets}
  \label{tab:slu}
  \centering
  \begin{tabular}{lccccc}
    \toprule
    System  	& \multicolumn{1}{c}{\bf{P1}} & \bf{E1} & \bf{E2} & \multicolumn{2}{c}{\bf{P2}} \\
    Beamwidth  & no & no & no & 1 & 4 \\ 
    \midrule

    \bf{FluentSpeech}	& 99.55 & 99.68 &  \bf{99.71} & 99.55 & \bf{99.68} \\
    \multicolumn{3}{l}{\bf{SmartLights}}   & &  &  \\
    close-field & 94.52 & 92.83 & \bf{95.18} & 94.88 & 95.12 \\
    far-field & 93.61 & 84.81 & 93.97 & 94.03 & \bf{94.22} \\
    \bottomrule
  \end{tabular}
\end{table}


\subsection{Decoding time}
\label{sec_cpu}

We then evaluated the decoding real-time factor (RTF) of all systems on
FluentSpeech, as presented in Table~\ref{tab:computation}. 

\begin{table}[htb]
  \caption{Real-time factors (RTF) of FluentSpeech decoded with different
  systems using an Intel Xeon\textregistered{} Gold 6130 CPU @ 2.10GHz}
  \label{tab:computation}
  \centering
  \begin{tabular}{lccccc}
    \toprule 
    System  	& \multicolumn{1}{c}{\bf{P1}} & \bf{E1} & \bf{E2} & \multicolumn{2}{c}{\bf{P2}} \\
    Beamwidth   & no  & no & no & 1 & 4 \\ 
    \midrule
	CPU & \bf{0.028} & 0.036 & \bf{0.028} & 0.047 & 0.140  \\
    \bottomrule
  \end{tabular}
\end{table}

System E2 is as efficient as P1, as its acoustic encoding
sequences are shortened to a length of approximately twice the number of
word-pieces in an utterance.

\section{Conclusion}

We compare four different SLU systems---both pipelines
and end-to-end systems---over two different SLU corpora. 
To rule out ASR biases and ensure a fair comparison, 
all use the same Conformer ASR model.
For the pipeline approach, better ASR 
accuracy naturally yields better SLU accuracy in general. 
Hence, the Conformer attention-rescoring decoding outperforms CTC decoding;
likewise, the pipeline system using attention-rescoring outperforms that using
CTC decoding. 
Impressively, the end-to-end E2 method beats
E1 in terms of both speed and accuracy.
Interestingly, in terms of SLU accuracy for E1, E2, and P1, 
E1 loses to P1 and P1 loses to E2.
There is no evidence for consistent two-stage superiority over end-to-end, 
or vice versa.

As far as the integration of ASR and NLU functions goes, 
current end-to-end SLU research is likely still at its early stage, 
and no approach has been developed that differs 
substantially from stacking two separate ASR and NLU modules together.
The authors propose a bold conjecture:
future SLU research might go even deeper in language modeling by
integrating an ASR word-association language model 
and an NLU word-to-intent language model.
Equipped with such a sophisticated language model,
the acoustic encoding model would be able to aggressively 
model more modalities of speech, emotions, and so on.

\bibliographystyle{IEEEtran}

\end{document}